\documentclass[10pt,conference]{IEEEtran}

\usepackage{epsfig}
\usepackage{graphicx}
\usepackage{latexsym}

\usepackage{bm}
\usepackage{amsmath}
\usepackage{amssymb}
\usepackage{amsfonts}

\usepackage{caption}
\usepackage{cite}
\usepackage{url}

\usepackage{multirow}

\usepackage{algorithm}
\usepackage[noend]{algpseudocode}

\DeclareMathOperator*{\argmin}{arg\,min}

\newcommand{\figg}{{Fig.~}}
\newcommand{\tabb}{{Table~}}

\newcommand{\algg}{{Algorithm~}}

\newenvironment{talign}
{\let\displaystyle\textstyle\align}
{\endalign}

\begin{document}

\title{Spatially-weighted Anomaly Detection \\ with Regression Mode}

\author{
\IEEEauthorblockN{Daiki Kimura$^{1}$, \:Minori Narita$^{2}$, \:Asim Munawar$^{1}$, \:Ryuki Tachibana$^{1}$}
	\IEEEauthorblockA{$^{1}$IBM Research AI, $^{2}$The University of Tokyo\\
		Email: \{daiki, asim, ryuki\}@jp.ibm.com, narita@g.ecc.u-tokyo.ac.jp}
}

\maketitle
\thispagestyle{empty}

\section*{Abstract}

Visual anomaly detection is common in several applications including medical screening and production quality check.
Although a definition of the anomaly is an unknown trend in data, in many cases some hints or samples of the anomaly class can be given in advance.
Conventional methods cannot use the available anomaly data, and also do not have a robustness of noise.
In this paper, we propose a novel spatially-weighted reconstruction-loss-based anomaly detection with a likelihood value from a regression model trained by all known data.
The spatial weights are calculated by a region of interest generated from employing visualization of the regression model.
We introduce some ways to combine with various strategies to propose a state-of-the-art method.
Comparing with other methods on three different datasets, we empirically verify the proposed method performs better than the others.

\section{Introduction}

Anomaly detection~\cite{chandola2009anomaly} has widely been used in many fields.
For example, it is now common to see the use for medical diagnosis~\cite{brainTumor2004,schlegl2017unsupervised}.
Depending on the real-world anomaly detection problems, some of the anomaly patterns might already be known.
For example, some typical different patterns from healthy older adults of a screening test for dementia called the Yamaguchi Fox-Pigeon Imitation Test~(YFPIT)~\cite{PSYG:PSYG373,yamaguchi} have been reported.
The detection should be able to leverage this information to improve the accuracy.
Since the manual check of anomaly conditions is mostly done by visual inspection and recent vision researches have led to significant breakthrough~\cite{resnet,kimura2011ultra}, using image information is natural for developing an automatic system~\cite{taboada2009anomaly}.
In this paper, we focus on visual anomaly detection for problems with some known anomalies.

A straightforward method where normal and anomaly patterns are given is to train a regression function for these classes by a convolutional neural network~(CNN).
However, this method suffers from the data imbalance problem, and the regression value for unknown patterns is intrinsically unexpected.
Hence, a structure for detecting anomaly pattern is required.

\begin{figure}[tb]
\begin{center}
\includegraphics[width=8.5cm]{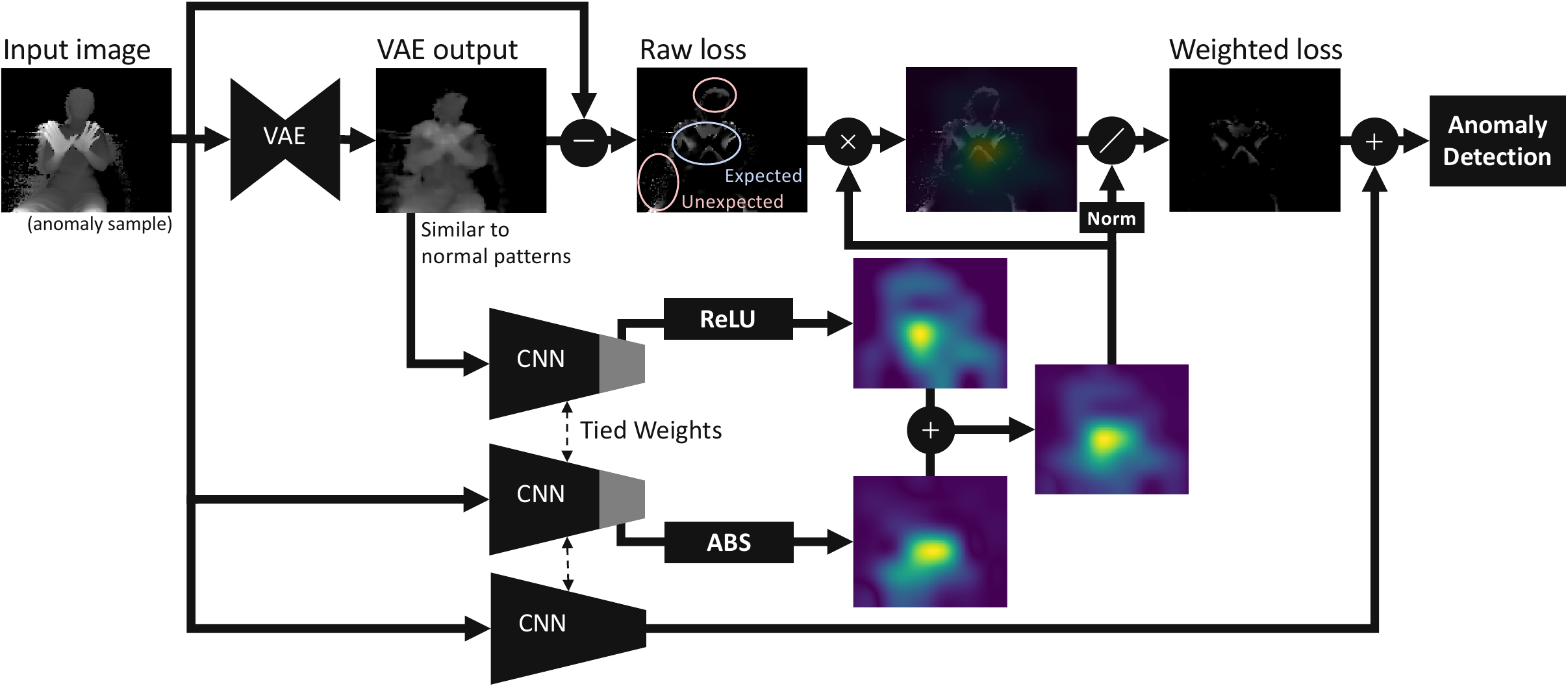}
\end{center}
\caption{Overview of the proposed method~(SPADER)}
\label{fig:spader}
\end{figure}

Visual anomaly detections~\cite{aead,vaead} mostly use a reconstruction-loss computed by a generative model which is minimized a loss between normal inputs and the reconstructed images.
However, noise in the image will be averaged or eliminated by the generative models; thus, these methods will misclassify noise as the anomaly.
The ``Raw loss'' image in \figg\ref{fig:spader} shows this issue; where the ``Unexpected'' loss is a problem.
Furthermore, these methods could not use the known anomaly patterns.

A region of interest~(ROI) can reduce this effect.
However, defining the ROI manually will be tricky, error-prone, and sub-optimal. 
Recently, Grad-CAM~\cite{gradcam} was proposed as a method which computes the ROI from gradients of CNN; it does not require any region information.

We propose a hybrid method of spatially-weighted reconstruction-loss-based anomaly detection and a likelihood value from a regression model trained known data.
The weights are computed by Grad-CAM~\cite{gradcam} to decrease noise effects, and a combination with the regression is to improve accuracy by known information.
Moreover, we introduce various strategies to combine.
We named the SPatially-weighted Anomaly DEtection as ``SPADE''~\cite{narita2018spatially}, and a method with SPADE and Regression as ``SPADER''. 
\figg\ref{fig:spader} shows a flow of the SPADER.
We verify the method with some methods on three datasets. 
The major contributions are, (1) we proposed a method for the condition where some anomaly patterns are known, (2) we proposed a spatially-weighted method for noise reduction, (3) we proposed a hybrid strategy to get state-of-the-art results.

\section{Proposed method}

As problem statement, there are three classes: normal class~$\mathbb{N}$, known anomaly class~$\mathbb{A}$, and unknown anomaly class~$\mathbb{U}$. 
Given a set of training patterns from $\mathbb{N}$ and $\mathbb{A}$, method does detection for test patterns of all classes.

\subsection{Training}

The proposed method have two training networks: a variational auto-encoder~(VAE)~\cite{vae} for reconstructing normal patterns, and a CNN for normalness regression.
 
The VAE network is set with the following equation,
\begin{equation}
\begin{split}
\{\bm{x_i}\} &\in \mathbb{N}, \\
\bm{\theta_v^ *} &= \argmin _{\bm{\theta_{v}}} \Big [- \sum_{i} \log p_{\bm{\theta_v}}(\bm{x_i}) \Big].
\end{split}
\end{equation}
\noindent During this minimization, the network optimizes,
\begin{equation}
\begin{split}
\mathcal { L } ( \bm{\theta} ,\bm{\phi} ; \bm{x_i} ) = &- D_{KL} \left( q _ { \bm{\phi} } ( \bm { z } | \bm{x_i} ) | | p _ { \bm{\theta} } ( \bm { z } ) \right) \\
&+ \mathbb { E } _ { q _ { \bm{\phi} } ( \bm{ z } | \bm { x_i } ) } \left[ \log p _ { \bm{\theta} } ( \bm{ x_i } | \bm{ z } ) \right],
\end{split}
\end{equation}
\noindent where $\bm{\theta}$ is the generative parameters, $\bm{\phi}$ is the variational parameters, and $\bm{z}$ is a random latent variable. 
In this paper, we use a normal distribution for the variable space; thus, the generative loss is a mean squared error.
In the remaining paper, we call $\bm{\theta_v}$ for both of $\bm{\theta}$ and $\bm{\phi}$.

The CNN network is set with the following equation,
\begin{equation}
\begin{split}
\{\bm{x_i}, y_i\} &\in \mathbb{N} \cup \mathbb{A}, \\
\bm{\theta_r^ *} &= \argmin_{\bm{\theta_r}} \sum_{i} \| y_i - f(\bm{x_i}; \bm{\theta_r})\|,
\end{split} 
\label{eq:cnn}
\end{equation}
\noindent where $y_i$ is the label value for $\bm{x_i}$.
When $\bm{x_i}$ is normal, $y_i = 1$; when $\bm{x_i}$ is anomaly, $y_i = 0$. 

\begin{algorithm}[tb]
\caption{SPADER}\label{euclid} 
\label{alg:proposed_method}
\begin{algorithmic}[1]
\State Given image $\bm{x}$, trained $\bm{\theta_r}$, and $\bm{\theta_v}$
\State $ \bm{\hat{x}} = g(\bm{x}; \bm{\theta_v})$ \Comment{Reconstructed image}
\State $ \bm{loss} = | \bm{\hat{x}} - \bm{x} | $ \Comment{Reconstruction-loss image}
\State $\bm{\alpha}_{\bm{x}} = \frac {1} {Z} \sum _ { i } \sum _ { j } \frac { \partial y ^ {1} _{\bm{x}} } { \partial A ^ { i j } _{\bm{x}} }$ 
\State $ \bm{cam}_{\bm{x}}^{\pm} = \text{Abs} \left( \sum_{k} \alpha^{k}_{\bm{x}} A^{k} \right)$ \Comment{Grad-CAM for input image}
\State $\bm{\alpha}_{\bm{\hat{x}}} = \frac {1} {Z} \sum _ { i } \sum _ { j } \frac { \partial y ^ {1} _{\bm{\hat{x}}} } { \partial A ^ { i j } _{\bm{\hat{x}}} }$ 
\State $ \bm{cam}_{\bm{\hat{x}}}^{+} = \text{ReLU} \left( \sum_{k} \alpha^{k}_{\bm{\hat{x}}} A^{k} \right)$ \Comment{Grad-CAM for VAE output}
\State $ \bm{cam} = \bm{cam}_{\bm{x}}^{\pm} + \bm{cam}_{\bm{\hat{x}}}^{+} $ 
\State $ \bm{loss^{sp}} = \frac {\bm{loss} * \bm{cam}} { \| \bm{cam} \|}$ \Comment{Spatially-weighted loss}
\State $ score = - \sum_{i} \sum_{j} loss^{sp}_{(i,j)} + f(\bm{x}; \bm{\theta_r})$
\Statex \Comment{Combine loss value and the likelihood}
\State Anomaly detection by $score$
\end{algorithmic}
\end{algorithm}

\subsection{Detection}

The proposed method has the following three steps to calculate a score for detection: getting the reconstruction-loss image, calculating an ROI image, and combining the spatially-weighted loss and a likelihood value from the regression model.
\algg\ref{alg:proposed_method} explains the details.

The reason for adding an ROI for the VAE output~(line~6-8 in \algg\ref{alg:proposed_method}) is, if we use only Grad-CAM of the input and the input is the anomaly, the area that is related to the normal class will not be focused on; however, the loss might appear in this region. 
Adding $\bm{cam}_{\bm{\hat{x}}}^{+}$ enables to focus on areas for the normal class as well.

The reason for using an absolute function~(line~5) is, the input is not only the normal class but also the anomaly class.
When we use a ReLU function that is same as the original Grad-CAM~\cite{gradcam}, ROI will only focus on the normal class.
However, as the reconstructed image by the VAE appears similar to the normal patterns, the function for VAE output~(line~7) must be a ReLU function.

The reason for adding normalization~(line~9) is, weighting without $\|\bm{cam}\|$ will be affected by the region size and strength of $\bm{cam}$.
When the ROI is wide or it has high value, $\bm{loss}*\bm{cam}$ will easily become a high value.
Therefore, we included the normalization in this equation.

\section{Experiments}


In this paper, we prepared the following three datasets that include noise: handwritten digit images~\cite{mnist} with noise, a public hand gesture dataset~\cite{hgk}, and images of human gestures which are described in~\cite{yamaguchi}.
The first and second datasets are for quantitative evaluation, and the third is for effectiveness assessment.

\subsection{Methods}

We compared the following methods in each dataset:
{
\setlength{\belowdisplayskip}{0pt}
\setlength{\abovedisplayskip}{0pt}
\begin{talign}
\intertext{{\bf VAE~\cite{vaead}}: VAE reconstruction-loss anomaly detection~\cite{vaead}}
score =& - \sum_{m}^{M} \sum_{i,j} | \bm{\hat{x}} - \bm{x} |
\intertext{{\bf Na\"ive VAE + GradCAM}: Na\"ive weighted-loss by ROI}
score =& - \sum_{m}^{M} \sum_{i,j} | \bm{\hat{x}} - \bm{x} | * \bm{cam}_{\bm{x}}^{+} \\
\intertext{{\bf SPADE w/o norm.}: SPADE without normalization}
score =& - \sum_{m}^{M} \sum_{i,j} | \bm{\hat{x}} - \bm{x} | * \left( \bm{cam_x^{\pm}} + \bm{cam_{\hat{x}}^{+}} \right) \\
\intertext{{\bf SPADE~\cite{narita2018spatially}}: Spatially-weighted anomaly detection}
score =& - \sum_{m}^{M} \sum_{i,j} \frac {\left| \bm{\hat{x}} - \bm{x} \right| * \left( \bm{cam_x^{\pm}} + \bm{cam_{\hat{x}}^{+}} \right)} { \left\| \bm{cam_x^{\pm}} + \bm{cam_{\hat{x}}^{+}} \right\|}
\intertext{{\bf CNN-Reg}: Anomaly detection by regression model}
score =& f(\bm{x}; \bm{\theta_r})
\intertext{{\bf VAE + CNN-Reg}: VAE-based detection w/ regression}
score =& - \frac{1}{M} \sum_{m}^{M} \sum_{i,j} | \bm{\hat{x}} - \bm{x} | + f(\bm{x}; \bm{\theta_r})
\intertext{{\bf SPADER: SPADE with regression}}
score =& - \frac{1}{M} \sum_{m}^{M} \sum_{i,j} \frac {\left| \bm{\hat{x}} - \bm{x} \right| * \left(\bm{cam_x^{\pm}} + \bm{cam_{\hat{x}}^{+}} \right)} { \left\| \bm{cam_x^{\pm}} + \bm{cam_{\hat{x}}^{+}} \right\|} + f(\bm{x}; \bm{\theta_r}) 
\end{talign}
}

\noindent where meanings of the variable are in \algg\ref{alg:proposed_method}, and $M$ is the number of trials for VAE reconstruction~($M = 5$). 

\subsection{MNIST with noise}

\begin{figure}[tb]
\begin{minipage}{0.42\linewidth}
\begin{center}
\includegraphics[width=3.5cm]{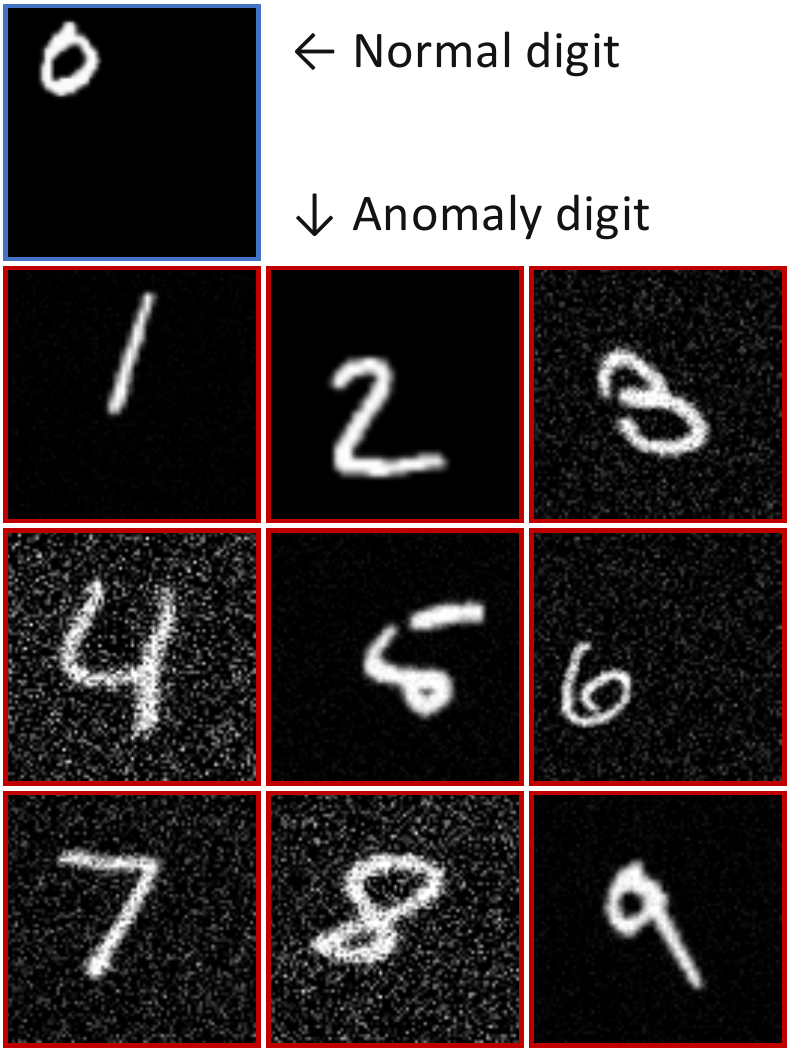}
\end{center}
\caption{Noisy MNIST (0: normal digit, \\1-9:~anomaly digit)}
\label{fig:mnist}
\end{minipage}
~
\begin{minipage}{0.57\linewidth}
\begin{center}
\includegraphics[width=4.8cm]{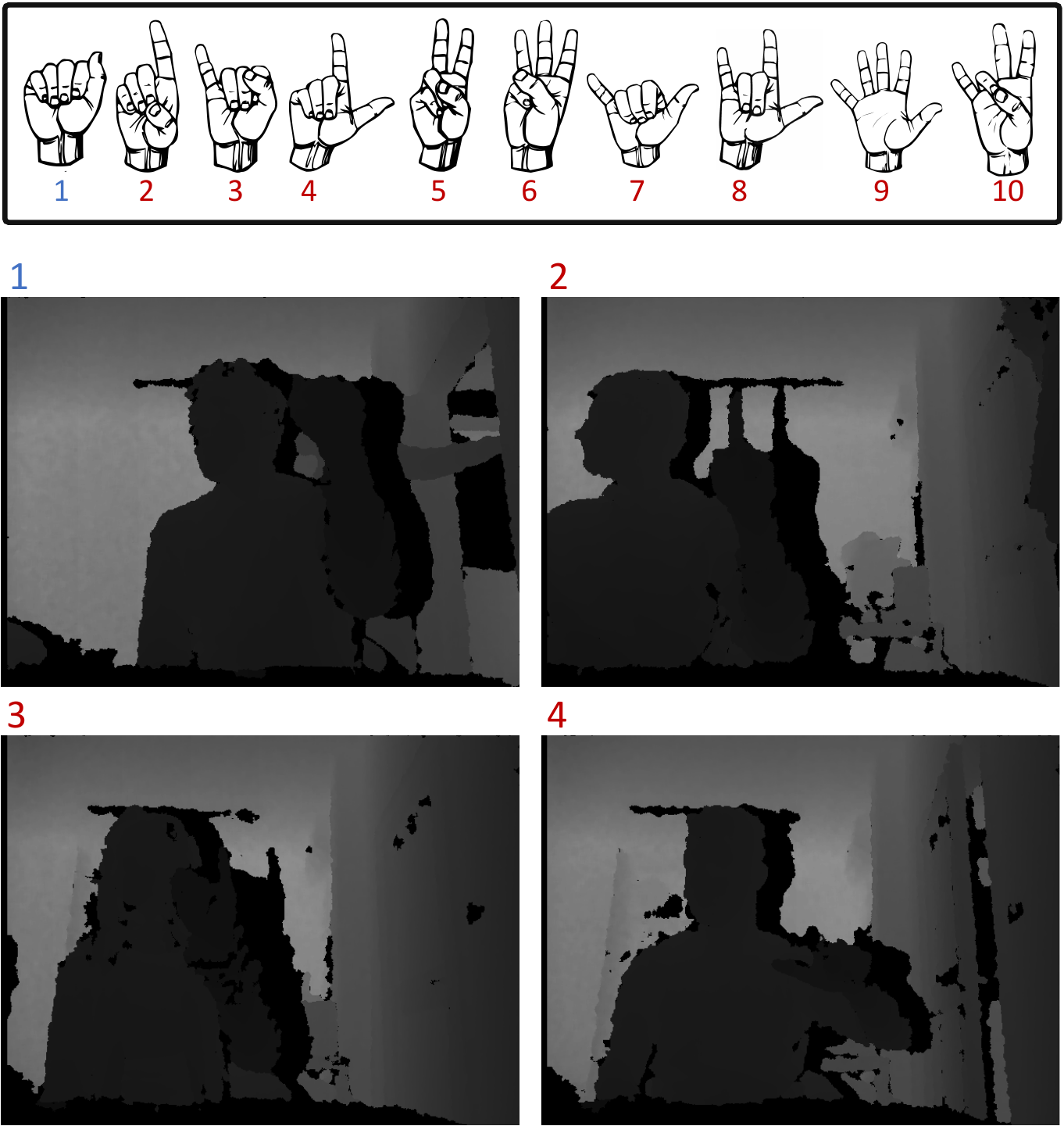}
\end{center}
\caption{Public hand gesture dataset~\cite{hgk} (top: class descriptions from \cite{hgk_dataset}, bottom: image examples)}
\label{fig:hgk}
\end{minipage}
\end{figure}

Since the MNIST~\cite{mnist} does not contain noise, the reconstruction-loss did not have a problem; An. J, {\it el al.} have already reported the performance~\cite{vaead}.
In this experiment, we added normal distribution noise~$\mathcal{N}(0,\sigma^2)$.
The $\sigma$ is generated from $\mathcal{N}(40,30^2)$ for each image.
This is because such a condition is common in a real case; for example, each image has different noise according to the difference of the person or background in \figg\ref{fig:hgk} and \figg\ref{fig:pigeon}.
We also changed the image size to $84 \times 84$, and the digit was put with random size and position.
\figg\ref{fig:mnist} shows the examples of generated images.
Here, `0' is the normal class, odd-numbers are candidates for the known anomaly class, and the others are the unknown anomaly.
The encoder and decoder of VAE have four convolutional~(conv) layers, and the latent space has 128 units.
The CNN for getting regression value and Grad-CAM has three conv layers.
Note that we did not change the network among the methods; we only changed the $score$ function. 

{
\renewcommand\arraystretch{1.2}
\begin{table*}[tb]
\begin{center}
\begin{tabular}{c|ccccc|c||ccc|c}
\hline
& \multicolumn{6}{c||}{Noisy MNIST~(original dataset \cite{mnist})} & \multicolumn{4}{c}{Pigeon~(the problem is from \cite{yamaguchi})} \\
\cline{2-11}
& \multicolumn{5}{c|}{Known anomaly digit} & \multirow{2}*{Average} & \multicolumn{3}{c|}{Known anomaly pose} & \multirow{2}*{Average} \\
& 1 & 3 & 5 & 7 & 9 & & c & d & e &\\
\hline
VAE~\cite{vaead} &    
.63$\pm$.01 &
.63$\pm$.01 &
.63$\pm$.01 &
.63$\pm$.01 &
.63$\pm$.01 &
.632±.01 &
.95$\pm$.01 &
.95$\pm$.01 &
.95$\pm$.01 &
.948±.01 \\
\hline
Na\"ive VAE + GradCAM &    
.67$\pm$.04 &
.59$\pm$.03 &
.65$\pm$.01 &
.76$\pm$.02 &
.53$\pm$.02 &
.640±.02 &
.80$\pm$.02 &
.71$\pm$.09 &
.78$\pm$.10 &
.762±.07 \\
SPADE w/o norm. &    
.85$\pm$.01 &
.83$\pm$.02 &
.88$\pm$.02 &
.87$\pm$.02 &
.85$\pm$.02 &
.857±.02 &
.96$\pm$.02 &
.96$\pm$.03 &
.82$\pm$.09 &
.911±.04 \\
SPADE~\cite{narita2018spatially} &    
.85$\pm$.04 &
.87$\pm$.01 &
.92$\pm$.02 &
.86$\pm$.04 &
.90$\pm$.03 &
.880±.03 &
.98$\pm$.01 &
.97$\pm$.01 &
.96$\pm$.03 &
.970±.02 \\
CNN-Reg~(Pigeon:\cite{resnet}) &    
.73$\pm$.04 &
.88$\pm$.02 &
.96$\pm$.02 &
.88$\pm$.02 &
.96$\pm$.01 &
.881±.02 &
.86$\pm$.03 &
.97$\pm$.02 &
.90$\pm$.03 &
.908±.03 \\
VAE + CNN-Reg &    
.73$\pm$.02 &
.81$\pm$.03 &
.87$\pm$.03 &
.85$\pm$.02 &
.89$\pm$.03 &
.831±.03 &
.97$\pm$.00 &
.99$\pm$.00 &
.98$\pm$.01 &
.980±.00 \\
{\bf Ours: SPADER} &    
{\bf .94$\pm$.02} &
{\bf .92$\pm$.01} &
{\bf .97$\pm$.00} &
{\bf .95$\pm$.01} &
{\bf .96$\pm$.01} &
{\bf .947±.01} &
{\bf .99$\pm$.00} &
{\bf 1.00$\pm$.00} &
{\bf .98$\pm$.01} &
{\bf .988±.01} \\
\hline
\end{tabular}
\end{center}
\caption{AUROC for noisy MNIST and pigeon dataset. Value is an average for 5 trials, and {\bf bold} is the best for each.}
\label{tab:result_mnist_pigeon}
\end{table*}
}

The left side of \tabb\ref{tab:result_mnist_pigeon} shows the averages of area under curve~(AUC) for ROC among 5 different trials with each condition.
For example, the second column shows the result for a condition that 0 is normal, 1 is known anomaly, and the others~(2-9) are unknown anomaly classes.
The proposed method~(SPADER) has the best performance.

\subsection{Hand gesture}

We used a public hand gesture dataset because we plan to apply gesture detection for YFPIT~\cite{yamaguchi} in the next section.
We used depth images in this dataset, the size of image is $640 \times 480$, and the images are taken from $14$~people with various backgrounds. 
\figg\ref{fig:hgk} explains class definitions and examples.
We set `1'-gesture as the normal, and the others are the anomaly class.
The encoder and decoder also have four conv layers, and the latent space has 256 units.
The CNN also consists three conv layers.

{
\renewcommand\arraystretch{1.2}
\begin{table*}[tb]
\begin{center}
\begin{tabular}{c|ccccccccc|c}
\hline
\multirow{2}*{Hand gesture~\cite{hgk}} & \multicolumn{9}{c|}{Known anomaly gesture} & \multirow{2}*{Average} \\
& 2 & 3 & 4 & 5 & 6 & 7 & 8 & 9 & 10 & \\
\hline
VAE~\cite{vaead} &    
.82$\pm$.17 &
.82$\pm$.17 &
.82$\pm$.17 &
.82$\pm$.17 &
.82$\pm$.17 &
.82$\pm$.17 &
.82$\pm$.17 &
.82$\pm$.17 &
.82$\pm$.17 &
.822$\pm$.17 \\
\hline
Na\"ive VAE + GradCAM &    
.82$\pm$.17 &
.80$\pm$.20 &
.77$\pm$.30 &
.78$\pm$.17 &
.76$\pm$.23 &
.80$\pm$.16 &
.80$\pm$.23 &
.81$\pm$.16 &
.77$\pm$.17 &
.790$\pm$.20 \\
SPADE w/o norm. &    
.83$\pm$.17 &
.81$\pm$.16 &
.87$\pm$.09 &
.83$\pm$.16 &
.80$\pm$.22 &
.83$\pm$.15 &
.84$\pm$.18 &
.82$\pm$.17 &
.82$\pm$.18 &
.828$\pm$.17 \\
SPADE~\cite{narita2018spatially} &    
.84$\pm$.18 &
.82$\pm$.16 &
.86$\pm$.13 &
.85$\pm$.15 &
.85$\pm$.16 &
.85$\pm$.15 &
.85$\pm$.16 &
.84$\pm$.18 &
.85$\pm$.16 &
.845$\pm$.16 \\
CNN-Reg &    
.77$\pm$.25 &
.94$\pm$.03 &
.92$\pm$.03 &
.96$\pm$.01 &
.95$\pm$.02 &
.87$\pm$.03 &
.94$\pm$.01 &
.85$\pm$.20 &
.97$\pm$.01 &
.908$\pm$.06 \\
VAE + CNN-Reg &    
.87$\pm$.20 &
{\bf .95$\pm$.04} &
.95$\pm$.04 &
{\bf .97$\pm$.04} &
.96$\pm$.03 &
.92$\pm$.05 &
.95$\pm$.04 &
.88$\pm$.20 &
.97$\pm$.02 &
.934$\pm$.07 \\
{\bf Ours: SPADER} &    
{\bf .87$\pm$.20} &
.95$\pm$.03 &
{\bf .95$\pm$.04} &
.96$\pm$.04 &
{\bf .96$\pm$.03} &
{\bf .92$\pm$.05} &
{\bf .95$\pm$.04} &
{\bf .88$\pm$.20} &
{\bf .97$\pm$.02} &
{\bf .937$\pm$.07} \\
\hline
\end{tabular}
\end{center}
\caption{AUROC for the hand gesture dataset~\cite{hgk}.  Value is an average for 5 trials, and {\bf bold} result is the best for each.}
\label{tab:result_hand}
\end{table*}
}

\tabb\ref{tab:result_hand} shows the AUC of the ROC curve for the detection. 
Sometimes the values of the proposed method are less than or similar to the `VAE + Reg' result; however, the SPADER shows the best results in total.

\subsection{Pigeon gesture}

\begin{figure}[tb]
\begin{center}
\includegraphics[width=8.4cm]{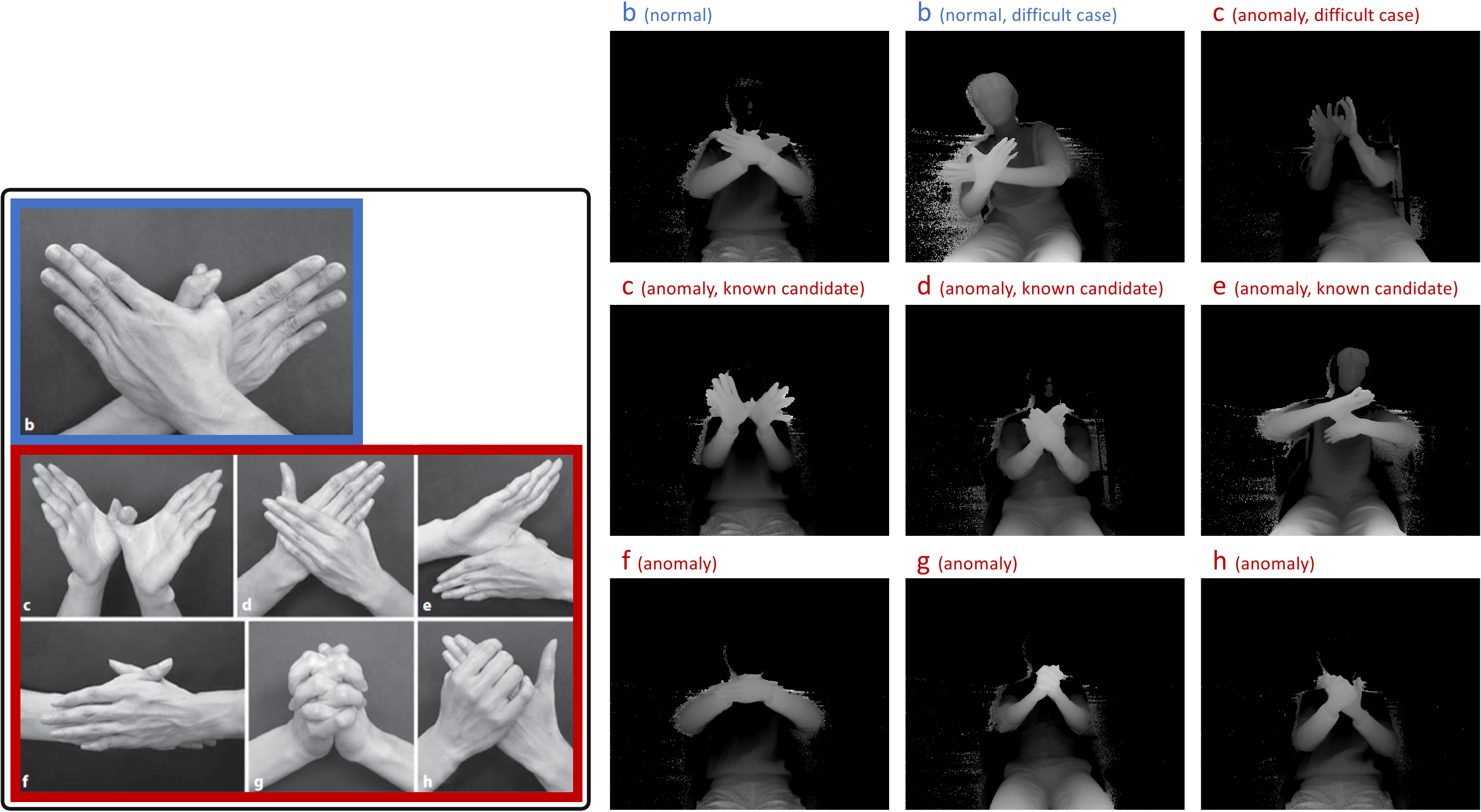}
\end{center}
\caption{Pigeon dataset~(left: class descriptions, the image is from \cite{yamaguchi}, right: samples of captured images for each class and some difficult cases)}
\label{fig:pigeon}
\end{figure}

We took images for the ``pigeon''-pose test of YFPIT~\cite{yamaguchi}.
We used a Kinect depth image, and we took from $18$~people.
During the shooting, we changed position and angle of the hand, close/open fingers, righty/lefty, sitting position, and stooping angle.
In total, we took around 189,000~images for 7~poses.
\figg\ref{fig:pigeon} shows the definition of poses, and the samples of took images and some difficult images. 
We set the `b'-pose as the normal class, `c, d, e' as candidates for the known anomaly class~(because Yamaguchi~{\it et~al.} reported these are typical patterns for the subjects), and the others as the unknown anomaly class. 
The encoder and decoder also have four conv layers, and the latent space also has 256 units.
The CNN has the ResNet~\cite{resnet} structure; however, the final layer is different.

The right side of \tabb\ref{tab:result_mnist_pigeon} shows the proposed method has the best results.
We hope this result helps to implement automatic YFPIT~\cite{yamaguchi}.
As future work, there are a comparison for different normal class, and a combination method with the latest visual explanation method~\cite{gradcamplusplus}.

\section{Conclusion}

We proposed a novel hybrid method with spatially-weighted anomaly detection and regression model.
We conducted experiments on three different datasets, then the proposed method produced the best performance compared to previous methods.
We hope this hybrid architecture will contribute in various applications including reinforcement learning works~\cite{kimura2018daqn}.

\bibliographystyle{IEEEtran}
\bibliography{ARXIV_SPADER}

\end{document}